\documentclass[10pt,twocolumn,letterpaper]{article}

\usepackage{cvpr}              

\usepackage{graphicx}
\usepackage{amsmath}
\usepackage{amssymb}
\usepackage{booktabs}

\usepackage[pagebackref,breaklinks,colorlinks]{hyperref}
\usepackage{multirow}

\usepackage[capitalize]{cleveref}
\crefname{section}{Sec.}{Secs.}
\Crefname{section}{Section}{Sections}
\Crefname{table}{Table}{Tables}
\crefname{table}{Tab.}{Tabs.}

\def\cvprPaperID{89} 
\def\confName{CVPR}
\def\confYear{2023}

\begin{document}

\title{ Hybrid Transformer and CNN Attention Network for \\ Stereo Image Super-resolution}

\author{
Ming Cheng$^{*1}$ \quad Haoyu Ma\thanks{Equal contribution.}\hspace{4pt}$^{1}$ \quad Qiufang Ma$^{1}$  \quad Xiaopeng Sun$^{1}$  \quad Weiqi Li$^{1, 2}$ \\
\quad Zhenyu Zhang$^{1, 2}$ \quad Xuhan Sheng$^{2}$ \quad Shijie Zhao\thanks{Corresponding author. (e-mail: zhaoshijie.0526@bytedance.com)}\hspace{4pt}$^{1}$ \quad Junlin Li$^{1}$ \quad Li Zhang$^{1}$\\
$^{1}$ByteDance Inc,\quad $^{2}$Peking University Shenzhen Graduate School\\
{\tt\small chengming.1129, mahaoyu.0510@bytedance.com}\\
}
\maketitle

\begin{abstract}
    %
    Multi-stage strategies are frequently employed in image restoration tasks. While transformer-based methods have exhibited high efficiency in single-image super-resolution tasks, they have not yet shown significant advantages over CNN-based methods in stereo super-resolution tasks. This can be attributed to two key factors: first, current single-image super-resolution transformers are unable to leverage the complementary stereo information during the process; second, the performance of transformers is typically reliant on sufficient data, which is absent in common stereo-image super-resolution algorithms. To address these issues, we propose a Hybrid Transformer and CNN Attention Network (HTCAN), which utilizes a transformer-based network for single-image enhancement and a CNN-based network for stereo information fusion. Furthermore, we employ a multi-patch training strategy and larger window sizes to activate more input pixels for super-resolution. We also revisit other advanced techniques, such as data augmentation, data ensemble, and model ensemble to reduce overfitting and data bias. Finally, our approach achieved a score of 23.90dB and emerged as the winner in Track 1 of the NTIRE 2023 Stereo Image Super-Resolution Challenge.

\end{abstract}

\section{Introduction}
\label{sec:intro}
Stereo image super-resolution aims to reconstruct high-resolution images from the given low-resolution left and right view images. It has attracted more attention in recent years as it has shown promising potential in downstream tasks such as stereo depth estimation and VR applications. Although stereo image super-resolution shares many similarities with single image super-resolution, there is a very important difference: single image super-resolution can only utilize the information from one view, while stereo image super-resolution can use the information from both views with a large overlapping area. This is not trivial because the information lost in one view might still exist in the other view, and utilizing the extra information from the other view can largely benefit the reconstruction process. Therefore, the final performance of the stereo image super-resolution algorithm largely relies on the feature extraction capability of each view, and the stereo information exchange capability.

The transformer architecture has proven to be very effective as a super-resolution algorithm in previous works~\cite{hat,swinir}. This is because a traditional convolutional neural network works on the locality prior and may suffer from long-range dependency. But transformers have a much larger receptive field compared to a convolutional neural network, and their self-attention mechanism can effectively model long-range dependencies. The effective feature extraction capability of a vision transformer can largely benefit stereo image super-resolution because the information from both views should be carefully utilized so that useful information is not lost during the super-resolution process.

On the other hand, the memory and computational cost for a vision transformer is usually much larger compared to a convolutional neural network. This problem will get much more severe when the resolution is high and a lot of query tokens exist. The long-range modeling capability also relies on a huge amount of training data. This makes the convolutional neural network performs well when the training data is limited. The previous state-of-the-art stereo super-resolution method NAFSSR~\cite{nafssr} is completely developed from a convolutional neural network, demonstrating the effectiveness of the convolutional neural network on the relatively small dataset. Because of its efficiency in computational cost, CNN-based models can usually afford more parallel exchange modules compared to transformer-based models and allows a more thorough information exchange.

Although there are many attempts to explore the possibility to combine both the convolutional neural network with a visual transformer~\cite{swinipassr} in the stereo image super-resolution, the optimal hybrid architecture still remains an open question. Based on the previous analysis, we propose a hybrid architecture that utilizes the strong long-range dependency modeling capability of a transformer and the effectiveness of information exchange between the two views of a convolutional neural network. In our proposed method, we utilize a transformer as our first stage to make sure most of the important features of the single view low-resolution images are kept for the further process, and a CNN-based method in the second stage to conduct effective stereo information exchange. The final performance illustrates the effectiveness of our design.
In summary, this paper has the following three contributions.
\begin{itemize}
    \item \textbf{A hybrid stereo image super-resolution network.} We propose a unified stereo image super-resolution algorithm, which integrates the transformer and CNN architectures, where the transformer is used to extract features of the single-view image while the CNN module is used to exchange the information from both views and generate the final super-resolution images.
    \item \textbf{Comprehensive data augmentation.} We conduct a comprehensive study on the multi-patch training strategy and other techniques and apply them to stereo image super-resolution.
    \item \textbf{A new state-of-the-art performance.} Our proposed method achieves new state-of-the-art performance and wins 1st place in Track 1 of the Stereo Image Super-resolution Challenge.
\end{itemize}

\section{Related Work}
\subsection{Single Image Super-Resolution}
The single image super-resolution is one of the most classic tasks in the field of low-level computer vision: the algorithm is expected to hallucinate a high-resolution image given a downsampled low-resolution image. This technique can benefit the algorithms in many other tasks such as tiny object detection \cite{tiny}, video super-resolution/enhancement \cite{video1,video2}, remote sensing \cite{remotesensing}, and blind image denoising \cite{gu2019blind}. At the very beginning, researchers utilize a database of external images or exemplars to generate the super-resolved image~\cite{timofte2013anchored,timofte2014a,yang2008image,zeyde2010single}. However, the performance of the hand-crafted features largely relies on the prior knowledge/hypothesis on the local image structure provided by the researchers. The local structures in images are usually very complicated, and these over-simplified hypotheses deteriorate the final performance of the hand-crafted features, which undermines the final performance of the super-resolution algorithm as these features are crucial in the process of retrieving the data from the external database.

\textbf{CNN-based methods.} By introducing the optimization algorithms such as gradient descent, the later introduced CNN-based methods relax the hand-crafted prior knowledge on the image structure and allow the network to learn the local structure pattern among a huge amount of data. It only keeps the locality prior of the image's local patterns to reduce the computational cost of the model. Many new devices are implemented to further improve the generalization capability of the model. SRCNN~\cite{dong2015image} introduces skip-connection in the process of super-resolution, which becomes a fundamental part of the following works. SRResNet~\cite{ledig2017photo} and EDSR~\cite{lim2017enhanced} use deeper and wider residual blocks and further improve the performance. RCAN~\cite{zhang2018image} incorporates the attention mechanism in a basic residual block to different channels and provides different emphasis on information from different channels. NAFNet~\cite{NAFNet} analyzes the previous works and reconstructs a simple baseline for image super-resolution which achieves state-of-the-art performance with reduced computational complexity.

\begin{figure*}
    \begin{center}
       \includegraphics[width=0.8\textwidth]{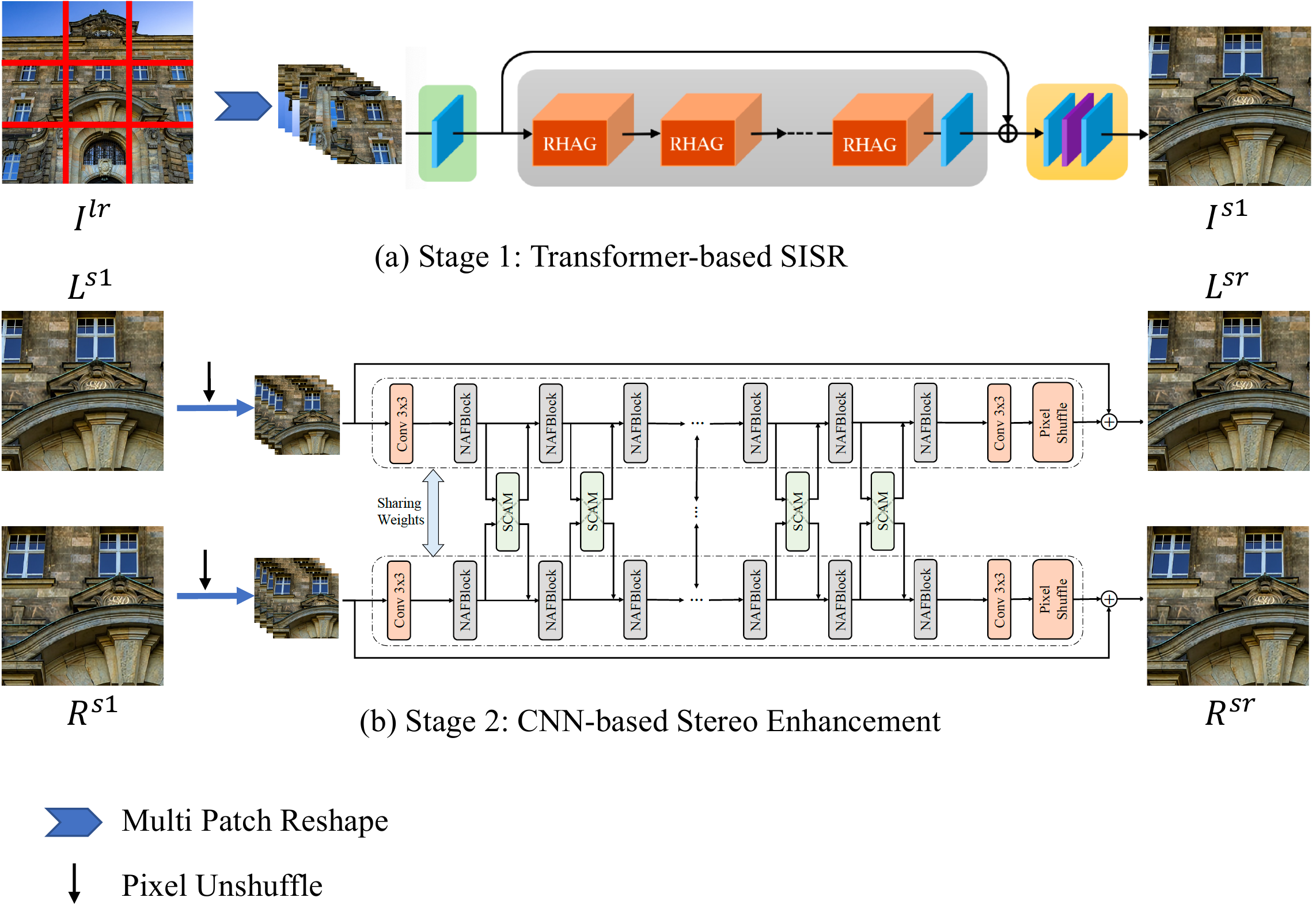}
    \end{center}
       \caption{Illustration of the proposed Hybrid Transformer and CNN Attention Network.}
    \label{fig:network}
\end{figure*}

\textbf{Transformer-based methods.} Transformers originated from the study of natural language processing and proves to be very effective in modeling long-range dependencies. It removes the prior knowledge about the locality which is used by a convolutional module and allows a much bigger receptive field. However, removing the prior knowledge also indicates more data is needed for the training so that the model can learn the prior knowledge during the training. IPT~\cite{IPT} evaluates the performance of a transformer structure in low-level tasks and illustrates that the simplest transformer can surpass the performance of a CNN given enough data. SwinIR~\cite{swinir} re-introduce the locality prior and use the windowed self-attention. HAT~\cite{hat} introduces the channel attention module which is firstly used in the CNN models and activates more pixels in its receptive field for the reconstruction. Both successful attempts in the SwinIR~\cite{swinir} and HAT~\cite{hat} prove that techniques in CNN-based methods can also benefit the transformer-based methods.

\subsection{Stereo Image Super-Resolution}
The major difference between single-image super-resolution and stereo-image super-resolution is that the stereo images may have redundant information in both views. Previous works are mostly developed from the single-image super-resolution backbone, and demonstrate the necessity of commnication modules between left and right views where super-resolution results from either view are usually much lower than the constructution results from both views~\cite{wang2022ntire}. 

Communication branches have been introduced in recent works to allow information exchange between the left and right views in stereo-image super-resolution~\cite{wang2019learning,ipassr,swinipassr,nafssr}. However, the disparity between the left and right views is typically along the epipolar line and larger than the receptive field of a traditional convolutional kernel. To address this, PASSR~\cite{wang2019learning} introduces a cross-attention module that allows for long-range dependency modeling. The attention mechanism is conducted only in the epipolar direction to reduce computational cost and memory consumption. The communication block is inserted in the middle of the network, and several losses are adopted simultaneously to achieve the best performance.
Building on the idea of PASSR, iPASSR~\cite{ipassr} introduces a biPAM module to aggregate information from both views after each residual block. It adopts a compact bi-directional parallax structure compared to its previous version in PASSR and effectively handles occlusions. Unlike PASSR and iPASSR, which use cross-attention mechanisms, SSRDE-FNet~\cite{dai2021feedback} explicitly models the disparity between the two views. It proposes a unified architecture to simultaneously estimate the view disparity and super-resolution results. The deep features from one view are warped according to the disparity and utilized to further improve the reconstructed SR result of the other view. However, a potential issue is the inconsistency of the reconstructed texture areas between the two views, which may lead to 3D fatigue for viewers.
To address this issue, SPAMNet~\cite{song2020stereoscopic} imposes a stereo-consistency constraint on the super-resolution results. Specifically, the reconstructed images from one view should be consistent when warped with the corresponding disparity map. A combination of self-attention and parallax attention is utilized to effectively aggregate information.

The rapid development of transformer-based super-resolution methods has led to increased interest in stereo-image super-resolution tasks. SwiniPASSR~\cite{swinipassr} explores the use of a parallax attention network for a transformer structure and conducts a progressive training strategy. Experimental results demonstrate that it can achieve much better results on the basis of a single-view transformer backbone. However, NAFSSR~\cite{nafssr} argues that a transformer network may not be necessary to achieve state-of-the-art performance. Instead, it simplifies the cross-attention module to allow it to be inserted after every block of the convolutional super-resolution block. This simplification allows for intensive information exchanges during the super-resolution process and greatly improves the final performance. This method achieves the new state-of-the-art performance and was the winner of the NTIRE 2022 competition~\cite{wang2022ntire}.

\section{Methods}

In this section, we present more details about the proposed Hybrid Transformer and CNN Attention Network (HTCAN). The proposed HTCAN is a multi-stage restoration network, as shown in Figure~\ref{fig:network}. 
More specifically, given the low-resolution stereo images $L^{lr}$ and $R^{lr}$, we first super-resolved them with a transformer-based single-image super-resolution network to $L^{s1}$ and $R^{s1}$.
In stage 2, we adopt a CNN-based network to stereo enhance $L^{s1}$ and $R^{s1}$ and get the enhanced images $L^{sr}$ and $R^{sr}$.
In the third stage, we use the same CNN-based network as stage 2 for further stereo enhancement and model ensembling.

\subsection{Stage 1: Transformer-based Single-image Super-resolution}
\textbf{Network Architecture.}
%
%
%
In the first stage, we adopt HAT-L~\cite{hat} as the backbone for single-image super-resolution (SISR) to super-resolve the input images $L^{lr}$ and $R^{lr}$.
%
To further increase the receptive field of HAT-L, we employ a multi-patch~\cite{mp} training strategy that unshuffle several neighboring patches into one patch. 
Then, the inputs of the SISR network are one low-resolution image patch and its eight surrounding patches, as shown in Figure~\ref{fig:network}(a).
The eight surrounding patches are cropped from the top, bottom, left and right of the center patch.
As a result, the eight surrounding patches may extend beyond the edge of the image. In such cases, we expand the image using reflect padding and extract the low-resolution patch and its eight surrounding patches from the padded image.
Given the nine input low-resolution patches, we first feed them into a $3\times 3$ convolutional layer to extract shallow features $F^1_L,F^1_R\in R^{H\times W\times C}$, where $C$ denotes the number of feature channels. In our experiments, we set $C$ to 180. 
The shallow features, which provide a preliminary perception of the inputs, are then fed into consecutive $K^1$ cascade Residual Hybrid Attention Group (RHAG)~\cite{hat} for self-attention and aggregate information.
We set $K^1$ as 12.
%
%
Furthermore, we increase the window size to $24\times 24$ for better information aggregation within windows.
Finally, after the efficient information aggregation by the cascade RHAGs, the super-resolved image is generated by convolution layers and pixel shuffle layers~\cite{shi2016real}.
The network output is the high-resolution patch corresponding to the center patch.
%

%
%

\textbf{Ensemble Strategy.}
We implemented self-ensembling by rotating and horizontally/vertically flipping the input low-resolution images. Additionally, we replaced the GeLU activation function in the HAT-L model with the SiLU activation function. Through experimentation, we found that the Fourier upsampling techniques introduced in \cite{fourier} did not significantly improve model performance. However, we discovered that introducing it as an additional ensemble model led to further performance improvements.

\subsection{Stage 2: CNN-based Stereo Enhancement}
\textbf{Network Architecture.}
The second stage aims to conduct stereo information exchange. To do this, we employ the state-of-the-art stereo super-resolution model NAFSSR-L~\cite{nafssr} as the backbone.
%
%
The NAFSSR-L is also a $4\times$ super-resolution model, while the upscaling is not necessary in this stage. 
As a result, we pixel unshuffle the input images from the stage 1 by 4 times to match the input-output dimension requirements of the second stage. The input channel of the first convolutional layer is also changed accordingly. 
In this way, we can reduce the memory occupancy and enlarge the receptive field of NAFSSR-L. We call this model UnshuffleNAFSSR-L.
We input the super-resolved images $L^{s1},R^{s1}$ from stage 1 into UnshuffleNAFSSR-L, as shown in Figure~\ref{fig:network}(b).
Given the unshuffled left and right images, we feed them into a $3\times 3$ convolutional layer, separately, to extract shallow features $F^2_L,F^2_R\in R^{H\times W\times C}$, where $C$ denotes the number of feature channels.
%
%
In our experiments, we set $C$ as 128.
Then, the shallow features are fed to consecutive $K^2$ cascade Nonlinear Activation Free (NAF) Blocks~\cite{nafssr} and Stereo Cross Attention Modules (SCAM) for cross-view information aggregation.
%
%
To ensure high-efficiency, the NAFBlocks replace traditional nonlinear activation functions with multiplication~\cite{vaswani2017attention}. 
We set $K^2$ to 128 in our experiments.
We insert one SCAM module between each two NAFblocks to enable cross-view information aggregation.
%
%
%
The SCAM module performs cross-attention on the left and right features based on Scaled Dot-Product Attention~\cite{vaswani2017attention}, which computes the dot products of the query with all keys and applies a softmax function to obtain the weights on the values. 
In the stereo image super-resolution task, the corresponding pixels between the left and right images are on the same horizontal line.
Thus, the SCAM module dot products all the tokens from the same horizontal lines in the left and right views, which captures the cross-view information in an efficient way.
After the high-efficient cross-view information aggragation by the cascade NAFBlocks and SCAMs, the stereo-enhanced images $L^{sr},R^{sr}$ are generated by convolution layers and pixel shuffle layers~\cite{shi2016real}, as shown in Figure~\ref{fig:network}(b).
%
%
%


\textbf{Ensemble strategy.}
We incorporate self-ensembling by horizontally and vertically flipping the input images and reversing the left and right views. To construct the final ensemble results, we select two models and average their outputs. It is important to note that we keep the outputs in float format to prevent any potential rounding errors.

\subsection{Stage 3: CNN-based Stereo Ensemble}
We notice that the ensemble output of the second stage is not satisfactory enough due to the lack of diversity of the models trained in the second stage. Therefore, we introduce the stage 3. The stage 3 is identical to the stage 2, except that the inputs are changed into the self-ensembled outputs from stage 2 instead of the counterparts from the stage 1. Although the performance of the model saturated in stage 3 and has no significant improvement comparing to stage 2, it serves as a good ensemble model and further improve the performance of the model trained in the stage 2. The overall performance changes across stages are illustrated as in Table \ref{tab:performance}. Due to time constraints, we only trained one stage 3 model.

\section{Experiments}
\subsection{Implementation Detail}

\textbf{Dataset.}
The Flickr1024 dataset \cite{wang2019flickr1024} is a widely used benchmark dataset for stereo image super-resolution tasks, consisting of 800 stereo image pairs for training, 112 pairs for validation, and 112 pairs for testing. The NTIRE 2023 Stereo Image Super-Resolution Challenge~\cite{Wang2023NTIRE} adopts the same training and validation settings as the Flickr1024 dataset, but includes an additional set of 100 stereo image pairs for testing.

\textbf{Training Details.}
Our proposed HTCAN network is a multi-stage network. Thus, we training our network in a multi-stage strategy.

\textit{Stage 1.}
In the training phase of stage 1, we train the network using Adam optimizer with $\beta_1=0.9$ and $\beta_2=0.99$. We first train the network with Charbonnier Loss and then finetune it with MSE loss. The model is trained for 800K iterations with mini-batches of size 32 and patch size $48\times 48$. 
To increase the diversity of training dataset for the stage 1, we use channel shuffle, horizontally and vertically flip, rotation and mixup for data augmentation.
The learning rate is initialized with $2e-4$ and reduced by half at [300K, 500K, 650K, 700K, 750K]. 
We implement our network with the Pytorch framework and train it using 8 NVIDIA Tesla A100 GPUs.

\textit{Stage 2.}
The training of the second stage model consists two phases. In the first phase, the NAFSSR model is trained with original code of NAFSSR-L on the Flickr1024 images. We use the default settings of the released code, and the result is about 0.04 dB lower than the result reported in the paper. This might because of the lack of training data. 

In the second phase, the model trained on the image are loaded as the pretrained model for the UnshuffleNAFSSR. We use the self-ensembled outputs in stage 1 as the inputs of stage 2 to train the model in this phase. We train the network using AdamW optimizer with $\beta_1=0.9$ and $\beta_2=0.9$. We first train the network with Charbonnier Loss and the finetune it with MSE loss. The model is trained for 300K iterations with mini-batches of size 32 and patch size $30\times 90$. 
To increase the diversity of training dataset for the stage 2, we use channel shuffle, horizontally and vertically flip for data augmentation.
We use TrueCosineAnnealingLR strategy to update the learning rate. The learning rate is initialized with $5e-4$ and the minimum learning rate is $1e-7$.  We implement our network with the Pytorch framework and train it using 8 NVIDIA Tesla A100 GPUs.

\textit{Stage 3.}
In the training phase of stage 3, we initialize the network with the pre-trained network in stage 2 and finetune the network with the output of stage 2.
Other training setting are the same as the stage 2.

\begin{table*}[!t]
\caption{Quantitative results achieved by different methods on the KITTI 2012~\cite{Geiger2012CVPR}, KITTI 2015~\cite{Menze2015ISA}, Middlebury~\cite{scharstein2014high}, and Flickr1024~\cite{wang2019learning} datasets. Here, PSNR/SSIM values achieved on both the left images (i.e., Left) and a pair of stereo images (i.e., (Left + Right) /2) are reported. The best results are in \textbf{bold faces}.}
\label{tab:compare}
  \renewcommand{\tabcolsep}{3pt}
  \renewcommand{\arraystretch}{1.0} 
\centering	
\begin{tabular}{lccccccc}
    \toprule
    \multirow{2}{*}{Methods} & \multicolumn{3}{c}{\textit{Left}} & \multicolumn{4}{c}{\textit{(Left+Right)/2}} \\
    \cmidrule(lr){2-4}
    \cmidrule(lr){5-8}
    & KITTI2012 & KITTI2015 & Middlebury & KITTI2012 & KITTI2015 & Middlebury & Flickr1024 \\
    \midrule
    VDSR~\cite{kim2016accurate} & 25.54/0.7662 & 24.68/0.7456 & 27.60/0.7933 & 25.60/0.7722 & 25.32/0.7703 & 27.69/0.7941 & 22.46/0.6718 \\
    EDSR~\cite{lim2017enhanced} & 26.26/0.7954 & 25.38/0.7811 & 29.15/0.8383 & 26.35/0.8015 & 26.04/0.8039 & 29.23/0.8397 & 23.46/0.7285 \\
    RDN~\cite{zhang2018residual} & 26.23/0.7952 & 25.37/0.7813 & 29.15/0.8387 & 26.32/0.8014 & 26.04/0.8043 & 29.27/0.8404 & 23.47/0.7295 \\
    RCAN~\cite{zhang2018image} & 26.36/0.7968 & 25.53/0.7836 & 29.20/0.8381 & 26.44/0.8029 & 26.22/0.8068 & 29.30/0.8397 & 23.48/0.7286 \\
    SwinIR~\cite{swinir} & 26.43/0.7996 & 25.60/0.7868 & 29.16/0.8379 & 26.52/0.8058 & 26.29/0.8098 & 29.25/0.8385 & 23.53/0.7322\\
    HAT-L~\cite{hat} & 26.91/0.8115 & 26.09/0.8014 & \textbf{30.53/0.8655} & 27.00/0.8177 & 26.83/0.8238 & \textbf{30.65/0.8672} & 24.21/0.7590\\
    \midrule
    StereoSR~\cite{jeon2018enhancing} & 24.49/0.7502 & 23.67/0.7273 & 27.70/0.8036 & 24.53/0.7555 & 24.21/0.7511 & 27.64/0.8022 & 21.70/0.6460 \\
    PASSRnet~\cite{wang2019learning} & 26.26/0.7919 & 25.41/0.7772 & 28.61/0.8232 & 26.34/0.7981 & 26.08/0.8002 & 28.72/0.8236 & 23.31/0.7195 \\
    SRRes+SAM~\cite{ying2020stereo} & 26.35/0.7957 & 25.55/0.7825 & 28.76/0.8287 & 26.44/0.8018 & 26.22/0.8054 & 28.83/0.8290 & 23.27/0.7233 \\
    IMSSRnet~\cite{lei2020deep} & 26.44/- & 25.59/- & 29.02/- & 26.43/- & 26.20/- & 29.02/- & -/- \\
    iPASSR~\cite{ipassr} & 26.47/0.7993 & 25.61/0.7850 & 29.07/0.8363 & 26.56/0.8053 & 26.32/0.8084 & 29.16/0.8367 & 23.44/0.7287 \\
    SSRDE-FNet~\cite{dai2021feedback} & 26.61/0.8028 & 25.74/0.7884 & 29.29/0.8407 & 26.70/0.8082 & 26.43/0.8118 & 29.38/0.8411 & 23.59/0.7352 \\
    Steformer~\cite{10016671} & 26.61/0.8037 & 25.74/0.7906 & 29.29/0.8424 & 26.70/0.8098 & 26.45/0.8134 & 29.38/0.8425 & 23.58/0.7376 \\
    NAFSSR-L~\cite{nafssr} & 27.04/0.8135 & 26.22/0.8034 & 30.11/0.8601 & 27.12/0.8194 & 26.96/0.8257 & 30.20/0.8605 & 24.17/0.7589 \\
    Ours & \textbf{27.16/0.8189} & \textbf{26.26/0.8083} & 30.25/0.8628 & \textbf{27.25/0.8249} & \textbf{26.99/0.8299} & 30.33/0.8634 & \textbf{24.44/0.7703}\\
    \bottomrule
\end{tabular}
\end{table*}
\begin{figure*}[h!]
	\footnotesize
	\centering
	\renewcommand{\tabcolsep}{1.0pt} 
	\renewcommand{\arraystretch}{0.7} 
	\begin{tabular}{ccccc}
	\multirow{2}{*}[9.5em]{\includegraphics[width=0.2\linewidth,height=0.329\linewidth]{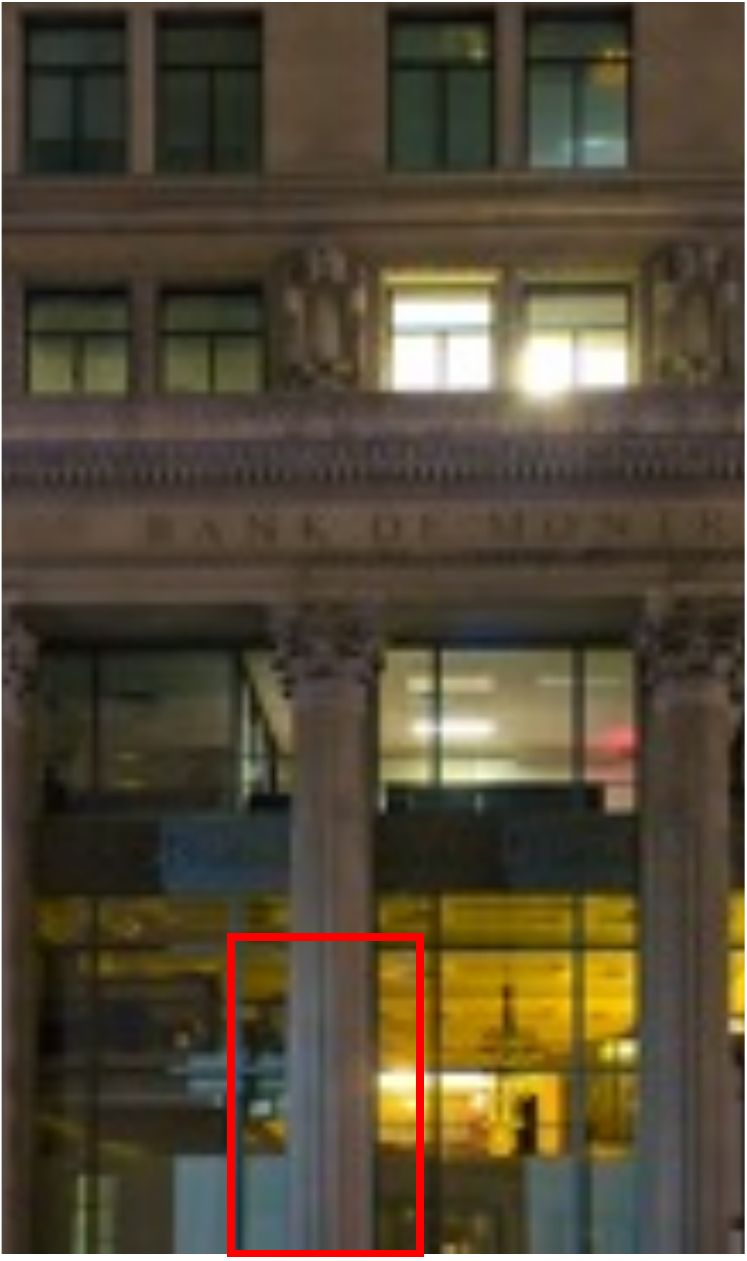}} &
        \includegraphics[width=0.14\linewidth,height=0.228\linewidth]{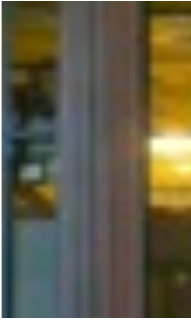} &
        \includegraphics[width=0.14\linewidth,height=0.228\linewidth]{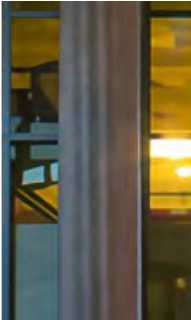} &
        \includegraphics[width=0.14\linewidth,height=0.228\linewidth]{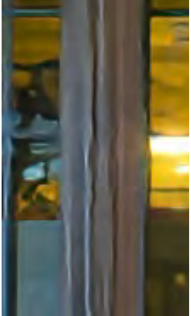} &
        \includegraphics[width=0.14\linewidth,height=0.228\linewidth]{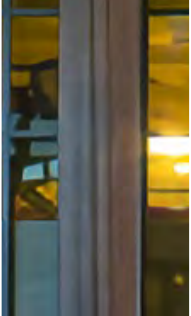} \\
        & (a) Bicubic & (b) HAT-L~\cite{hat} & (c) StereoSR~\cite{jeon2018enhancing} & (d) iPASSR~\cite{ipassr} \\
        &\includegraphics[width=0.14\linewidth,height=0.228\linewidth]{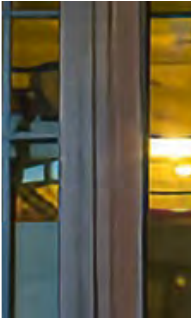} &
        \includegraphics[width=0.14\linewidth,height=0.228\linewidth]{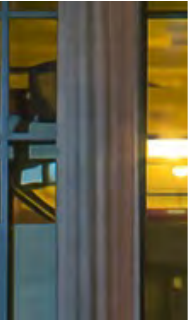} &
        \includegraphics[width=0.14\linewidth,height=0.228\linewidth]{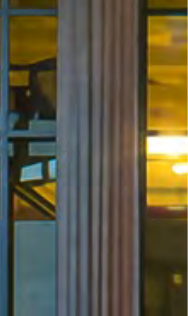} &
        \includegraphics[width=0.14\linewidth,height=0.228\linewidth]{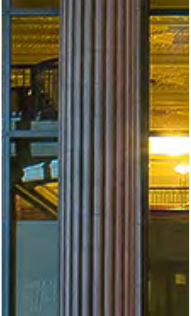} \\
        & (e) SRRes+SAM~\cite{ying2020stereo} & (f) NAFSSR-L~\cite{nafssr} & Ours & GT \\
        \multirow{2}{*}[9.5em]{\includegraphics[width=0.2\linewidth,height=0.329\linewidth]{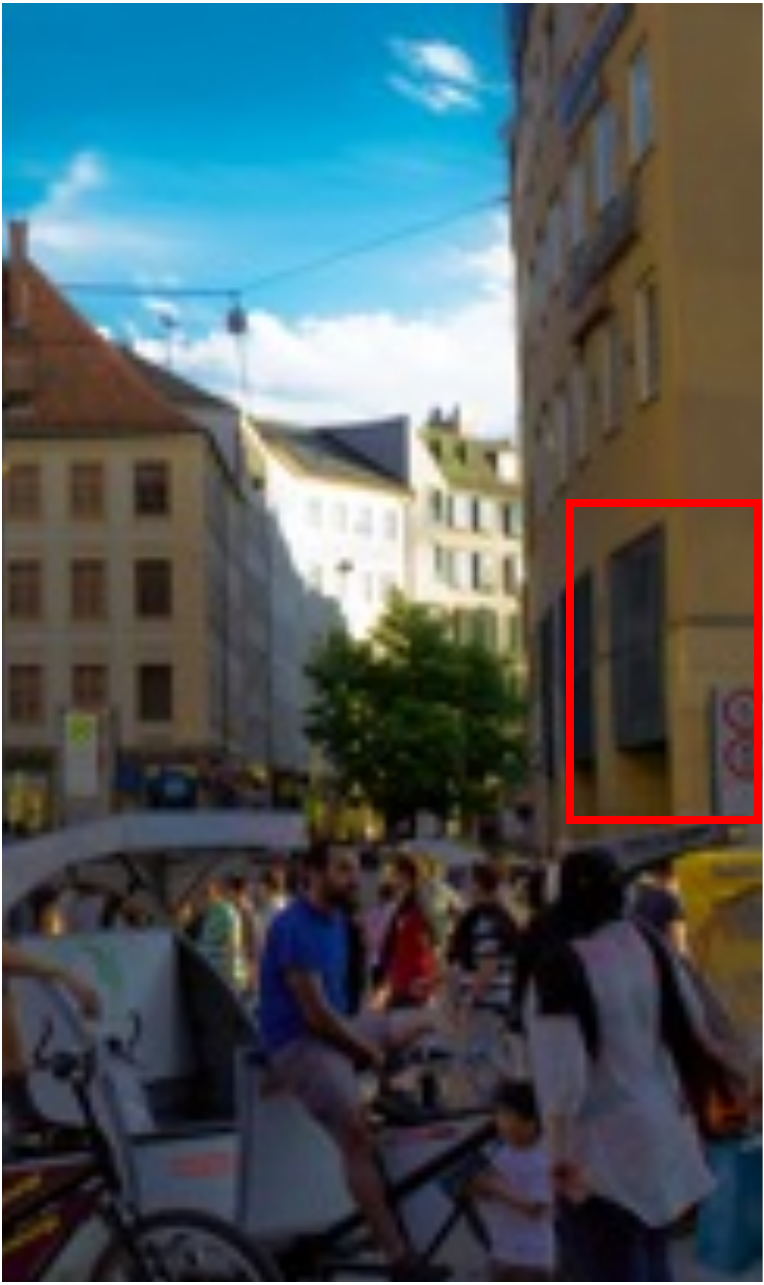}} &
        \includegraphics[width=0.14\linewidth,height=0.228\linewidth]{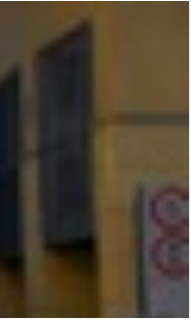} &
        \includegraphics[width=0.14\linewidth,height=0.228\linewidth]{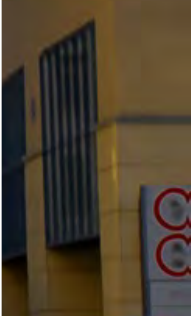} &
        \includegraphics[width=0.14\linewidth,height=0.228\linewidth]{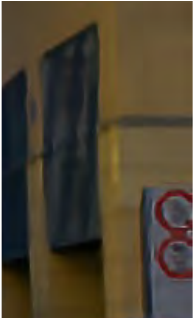} &
        \includegraphics[width=0.14\linewidth,height=0.228\linewidth]{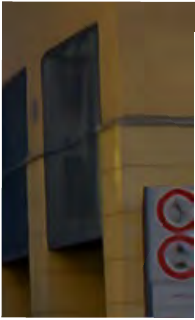} \\
        & (a) Bicubic & (b) HAT-L~\cite{hat} & (c) StereoSR~\cite{jeon2018enhancing} & (d) iPASSR~\cite{ipassr} \\
        &\includegraphics[width=0.14\linewidth,height=0.228\linewidth]{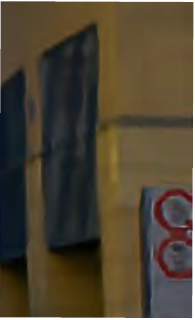} &
        \includegraphics[width=0.14\linewidth,height=0.228\linewidth]{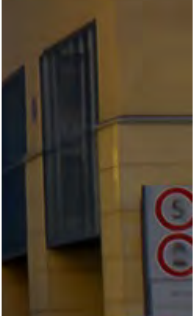} &
        \includegraphics[width=0.14\linewidth,height=0.228\linewidth]{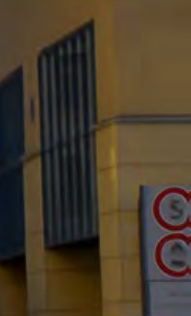} &
        \includegraphics[width=0.14\linewidth,height=0.228\linewidth]{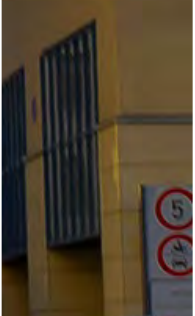} \\
        & (e) SRRes+SAM~\cite{ying2020stereo} & (f) NAFSSR-L~\cite{nafssr} & Ours & GT \\
        \end{tabular}
	\caption{{\textbf{Visual comparisons on Flickr1024 dataset~\cite{wang2019learning}.} GT denotes the ground truth.} }
	\label{fig:sota}
\end{figure*}

\textbf{Comparison to State-of-the-arts Methods.}
We conducted a comprehensive comparison of our method with state-of-the-art single-image super-resolution methods, namely VDSR~\cite{kim2016accurate}, EDSR~\cite{lim2017enhanced}, RDN~\cite{zhang2018residual}, RCAN~\cite{zhang2018image}, SwinIR~\cite{swinir}, HAT-L~\cite{hat}, and stereo-image super-resolution methods, including StereoSR~\cite{jeon2018enhancing}, PASSRnet~\cite{wang2019learning}, SRRes+SAM~\cite{ying2020stereo}, IMSSRnet~\cite{lei2020deep}, iPASSR~\cite{ipassr}, SSRDE-FNet~\cite{dai2021feedback}, NAFSSR-L~\cite{nafssr}, Steformer~\cite{10016671}.
It is worth mentioning that NAFSSR-L~\cite{nafssr} is trained on Flickr1024~\cite{wang2019learning} and Middlebury~\cite{scharstein2014high} datasets, whereas HAT-L is pre-trained on ImageNet~\cite{deng2009imagenet} dataset and then fine-tuned on the DIV2K~\cite{lim2017enhanced}+Flicker2K~\cite{timofte2017ntire} datasets.
In contrast, our model is only trained on the Flickr1024~\cite{wang2019learning} dataset provided by the NTIRE 2023 Stereo Image Super-Resolution Challenge~\cite{Wang2023NTIRE}.
Our method, presented in Table~\ref{tab:compare}, includes self-ensemble and model ensemble strategies.
 
To evaluate the performance of our method, we utilized 20 images from KITTI 2012~\cite{Geiger2012CVPR}, 20 images from KITTI 2015~\cite{Menze2015ISA}, 5 images from Middlebury~\cite{scharstein2014high}, and 112 images from the test set of Flickr1024~\cite{wang2019learning} for testing, as shown in Table~\ref{tab:compare}. Note that, unlike Table~\ref{tab:rank}, the Flickr1024 test images used in Table~\ref{tab:compare} are from the test set of the original Flickr1024~\cite{wang2019learning} dataset not the test set from the NTIRE 2023 Stereo Image Super-Resolution Challenge~\cite{Wang2023NTIRE}. We followed the evaluation protocol of iPASSR~\cite{ipassr} and report PSNR/SSIM scores on the left images with their left boundaries (64 pixels) cropped, and average scores on stereo image pairs (i.e., (Left + Right) /2) without any boundary cropping. The results are summarized in Table~\ref{tab:compare}. 

\begin{table}[t]
	\centering
	\caption{Performance improvements across stages (all PSNR values are calculated on the validation set. EM$\times 8$ denotes self-ensemble $\times 8$.)}
        \renewcommand{\tabcolsep}{3.0pt}
        \renewcommand{\arraystretch}{1.0} 
	\begin{tabular}{l c c}
		\toprule
		Model & PSNR & PSNR(EM$\times 8$) \\
		\midrule
            Stage 1: HAT-Charbonier w/o DA & 23.83 &\\
            Stage 1: HAT-Charbonier & 23.90 &\\
		Stage 1: MPHAT-Charbonier & 23.94 &\\
            Stage 1: MPHAT-MSE & 23.98 & 24.04\\ 
            Stage 1: MPHAT-SiLU & 23.97 & 24.04\\
            Stage 1: MPHAT-FourierUp & 23.97 & 24.04\\
            Stage 1: MPHAT-WindowSize24 &  23.98 & 24.05\\
            Stage 1: MPHAT-Ensemble & 24.07 & \\
            \midrule
            Stage 2: UnshuffleNAFSSR-L & 24.34 & 24.34\\
            Stage 2 $\&$ Stage 3 Ensemble & 24.34 & \\
		\bottomrule
	\end{tabular}
	\label{tab:performance}
\end{table}

\subsection{Results}
As shown in Table~\ref{tab:compare} and Figure~\ref{fig:sota}, our method outperfroms other state-of-the-art single-image super-resolution methods and stereo-image super-resolution methods on most test datasets.
On the in-distribution dataset Flickr1024~\cite{wang2019learning}, our method achieved a 0.27dB gain over NAFSSR-L\cite{nafssr} and a 0.23dB gain over HAT-L~\cite{hat}.
On the out-distribution dataset Middlebury~\cite{scharstein2014high}, HAT-L~\cite{hat} achieved the best results, likely due to its pre-training on the large-scale ImageNet~\cite{deng2009imagenet} dataset.
From the visual results in Figure~\ref{fig:sota}, the reconstructed pillars of NAFSSR-L~\cite{nafssr} and HAT-L~\cite{hat} appear blurred.
However, the reconstructed pillars of our method is much sharper and more textured.
While HAT-L~\cite{hat} can reconstruct the fine-grained textures of the window, it misses the number "5" on the guidepost. Conversely, NAFSSR-L~\cite{nafssr} can restore the number "5" on the guidepost, but misses the textures on the window. In contrast, our method reconstructed both the fine-grained textures of the window and the number, achieving superior visual quality.

\begin{figure*}[ht]
	\footnotesize
	\centering
	\renewcommand{\tabcolsep}{1.0pt} 
	\renewcommand{\arraystretch}{0.7} 
	\begin{tabular}{cccc}
	\includegraphics[width=0.23\linewidth]{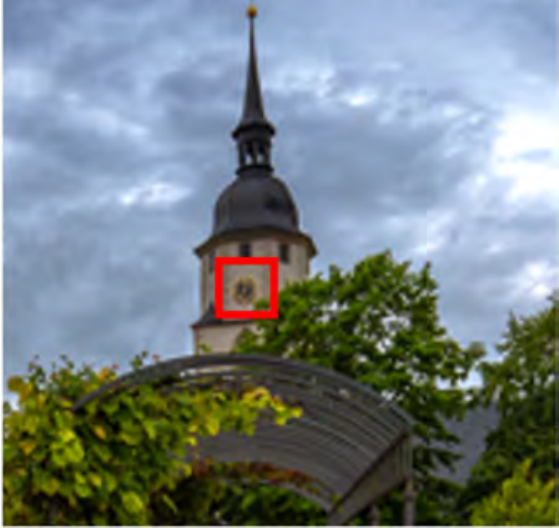} &
        \includegraphics[width=0.23\linewidth]{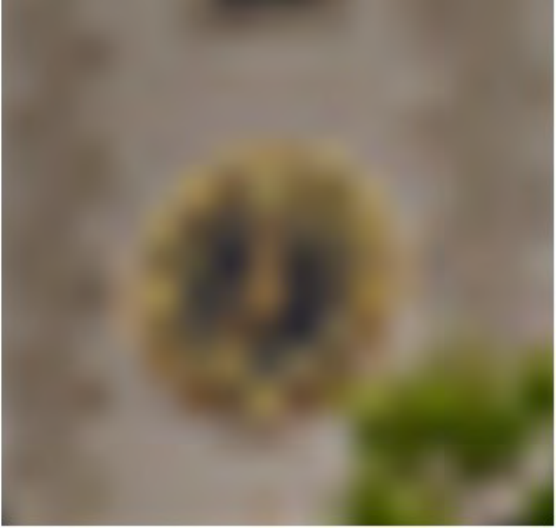} &
        \includegraphics[width=0.23\linewidth]{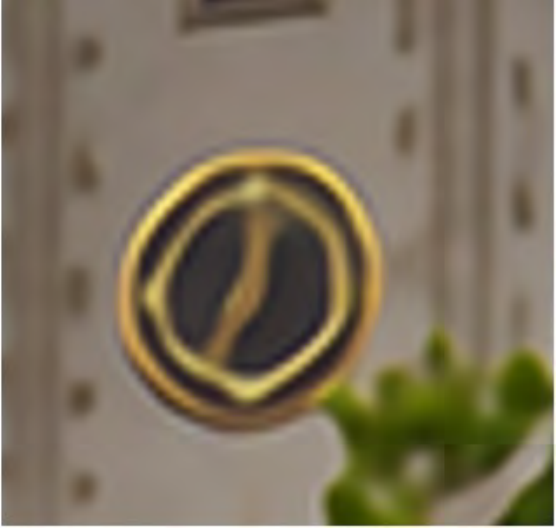} &
        \includegraphics[width=0.23\linewidth]{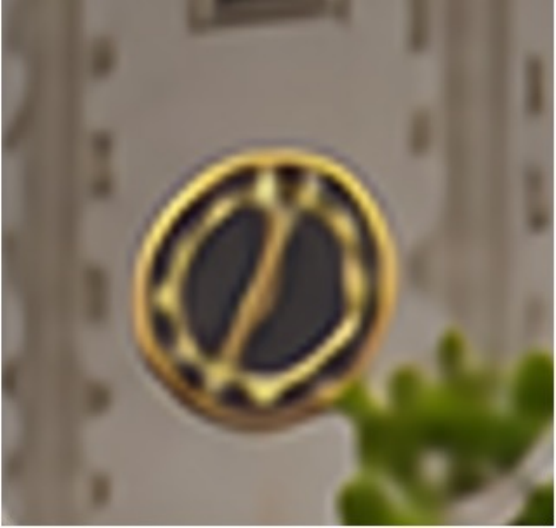} \\
        \includegraphics[width=0.23\linewidth]{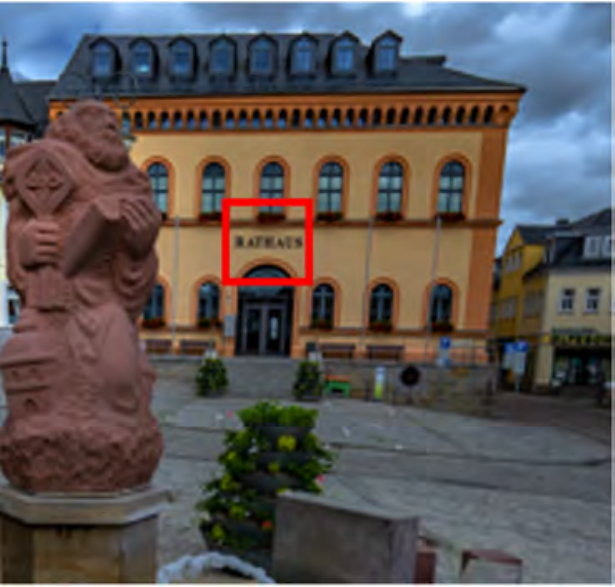} &
        \includegraphics[width=0.23\linewidth]{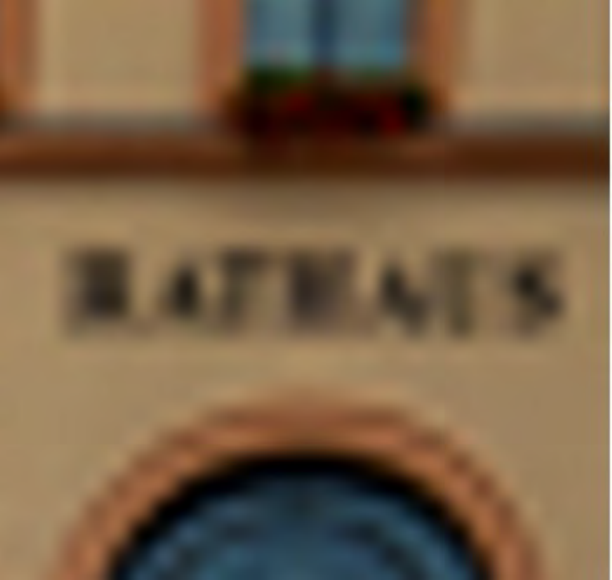} &
        \includegraphics[width=0.23\linewidth]{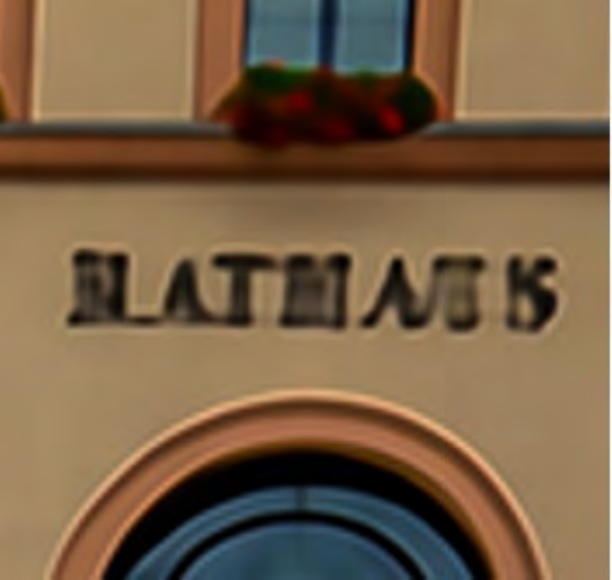} &
        \includegraphics[width=0.23\linewidth]{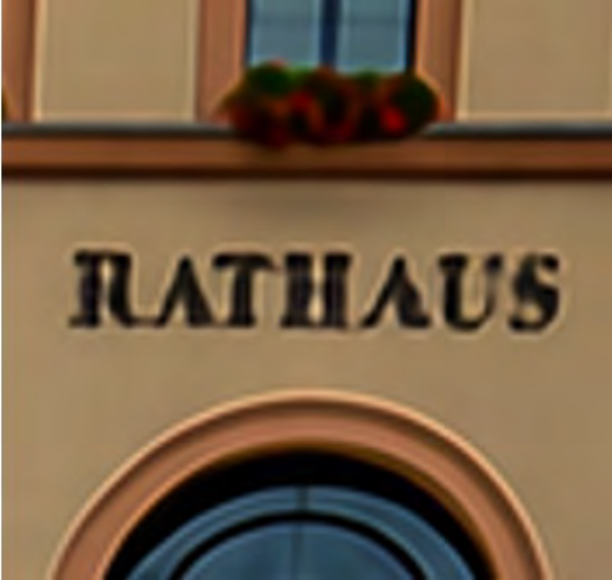} \\
        \includegraphics[width=0.23\linewidth]{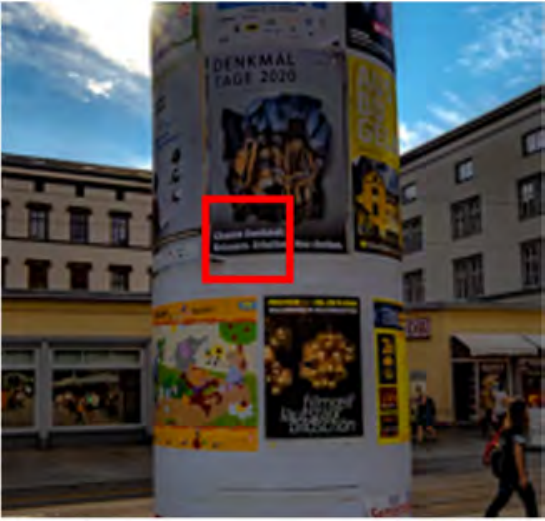} &
        \includegraphics[width=0.23\linewidth]{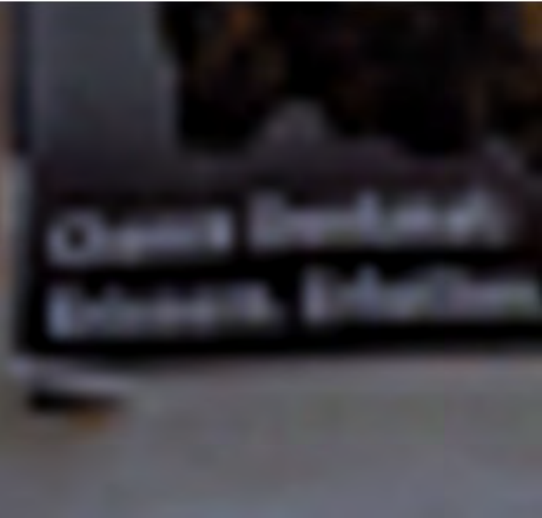} &
        \includegraphics[width=0.23\linewidth]{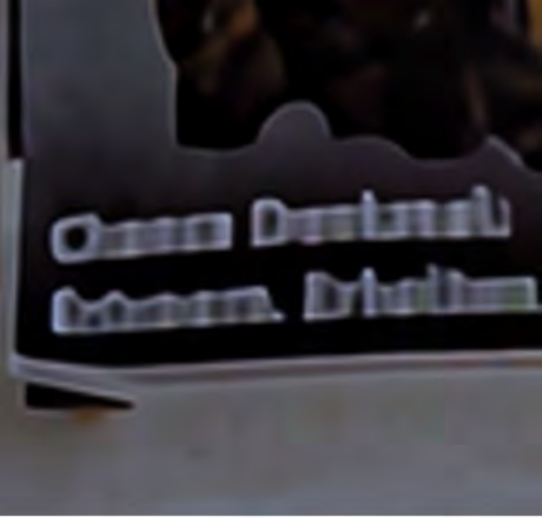} &
        \includegraphics[width=0.23\linewidth]{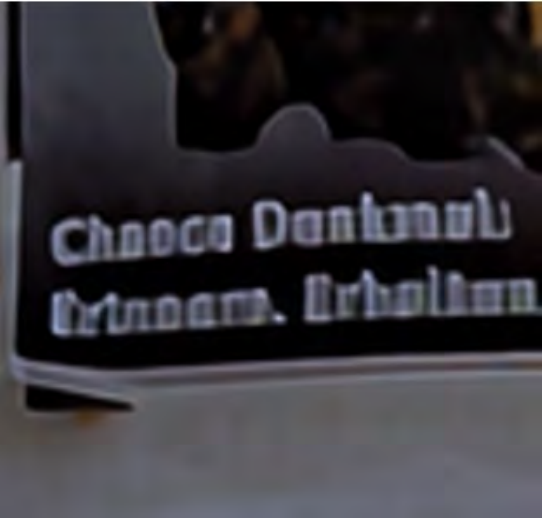} \\
         & (a) Bicubic & (b) Transformer-based SISR & (c) CNN-based stereo enhancement \\
        \end{tabular}
	\caption{\textbf{Visual analyses on the proposed multi-stage strategy.} }
	\label{fig:results}
\end{figure*}

\subsection{Analyses}
We analyze the proposed model with the validation set provided by the NTIRE 2023 Stereo Image Super-Resolution Challenge~\cite{Wang2023NTIRE}, as shown in Table~\ref{tab:performance}.
%
%
In this study, we conducted several experiments to improve the performance of our proposed method for stereo-image super-resolution, as shown in Table~\ref{tab:performance}.
%
First, we introduced the data augmentation methods channel shuffle and mixup, denoted by the abbreviation "DA," which resulted in a performance gain of 0.07 dB. These methods have been shown to be effective in improving the generalization ability of deep neural networks by introducing random perturbations to the input data.
The abbreviation "MPHAT" denotes HAT-L with multi-patch training strategy. The proposed multi-patch training strategy improves performance by 0.04 dB. 
The reason is that the multi-patch training strategy expand the receptive field of the network, thus improve the information aggregation within single image.
We further improved the performance by finetuning the MPHAT model with mean squared error (MSE) loss, which resulted in an additional performance gain of 0.04 dB.
%
Additionally, we found that using a larger window size ($24\times 24$) resulted in a 0.01 dB gain. This strategy involves using a larger receptive field to capture more contextual information and improve the network's ability to reconstruct fine-grained details.
%
To further improve our method's performance, we adopted a self-ensemble strategy, which improved the performance by 0.07 dB.
We rotate and horizontally/vertically flip the input image and averaging the results to obtain a final prediction, which has been shown to be effective in reducing the variance of the network's output.
%
%
Finally, we presented the results of the model ensemble, denoted by "MPHAT-Ensemble," which combined the results of MPHAT-MSE (EM$\times 8$), MPHAT-SiLU (EM$\times 8$), MPHAT-FourierUP (EM$\times 8$), and MPHAT-WindowSize24 (EM$\times 8$), weighted according to their respective performances. Specifically, we weighted the results of these model with weights of 1/7, 1/7, 1/7, and 4/7, respectively.
%
%
We set the largest weight on MPHAT-WindowSize24 because it outperforms the other three models. The model ensemble strategy further improves performance by 0.02 dB.
%
In our proposed method for stereo-image super-resolution, we introduced a CNN-based stereo enhancement module to further improve the performance. This module leverages the cross-view information from both the left and right stereo images to enhance the details and features of the reconstructed image. As we reported in Table~\ref{tab:performance}, this module significantly boosted the performance by 0.27 dB.
%

As shown in Figure~\ref{fig:results}, our proposed multi-stage method combines transformer-based single-image super-resolution (SISR) and CNN-based stereo enhancement to recover fine-grained spatial details. The transformer-based SISR method is used to upscale the low-resolution input images to a higher resolution, while the CNN-based stereo enhancement module is used to enhance the details by leveraging the cross-view information. 
Please look at the clock on the tower and the text on the building. Our proposed multi-stage strategy has effectively restored the textures.

\begin{table}[t]
	\centering
	\caption{Final results of ``NTIRE 2023 Challenges on Stereo Image Super-resolution - Track 1".)}
        \renewcommand{\tabcolsep}{13.0pt}
        \renewcommand{\arraystretch}{1.0} 
	\begin{tabular}{c c }
		\toprule
		Rank & PSNR \\
		\midrule
		\textbf{1st (Ours)} & \textbf{23.8961} \\
            2nd & 23.8911 \\
            3rd & 23.8830\\
            4th & 23.8220 \\
            5th & 23.8041 \\
             6th & 23.7719 \\
              7th & 23.7560 \\
               8th & 23.7424 \\
               9th & 23.7252 \\
               10th & 23.7181 \\
		\bottomrule
	\end{tabular}
	\label{tab:rank}
\end{table}

\subsection{NTIRE 2023 Challenge}
%
%
The primary objective of the NTIRE 2023 Stereo Image Super-resolution Challenge~\cite{Wang2023NTIRE} was to obtain a solution that can produce sharp results with high fidelity (PSNR) from bicubically downsampled input images to their corresponding ground truth images. Our proposed  hybrid Transformer and CNN Attention Network achieved first place in Track 1 of the challenge, as shown in Table~\ref{tab:rank}. We would like to note that our method utilized self-ensemble and model ensemble strategies to improve its performance. Ultimately, our method achieved a PSNR of 24.34 dB on the validation set and 23.90 dB on the test set, as presented in Tables~\ref{tab:performance} and~\ref{tab:rank}.

\section{Conclusions}

In this paper, we introduce the Hybrid Transformer and CNN Attention Network (HTCAN), which employs a two-stage approach to super-resolve low-resolution stereo images using a transformer-based SISR module and a CNN-based stereo enhancement module. Our proposed multi-patch training strategy and large window size increase the number of input pixels activated during the SISR phase, resulting in a 0.05dB gain over the original HAT-L~\cite{hat} architecture. Furthermore, our method employs advanced techniques, including data augmentation, data ensemble, and model ensemble, to achieve a PSNR of 23.90dB on the test set and win the 1st place in Track 1 of the Stereo Image Super-resolution Challenge.

{\small
\bibliographystyle{ieee_fullname}
\bibliography{egbib}

\begin{thebibliography}{10}\itemsep=-1pt

\bibitem{tiny}
Yancheng Bai, Yongqiang Zhang, Mingli Ding, and Bernard Ghanem.
\newblock Finding tiny faces in the wild with generative adversarial network.
\newblock In {\em Proceedings of the IEEE conference on computer vision and
  pattern recognition}, pages 21--30, 2018.

\bibitem{IPT}
Hanting Chen, Yunhe Wang, Tianyu Guo, Chang Xu, Yiping Deng, Zhenhua Liu, Siwei
  Ma, Chunjing Xu, Chao Xu, and Wen Gao.
\newblock Pre-trained image processing transformer.
\newblock In {\em Proceedings of the IEEE/CVF Conference on Computer Vision and
  Pattern Recognition}, pages 12299--12310, 2021.

\bibitem{NAFNet}
Liangyu Chen, Xiaojie Chu, Xiangyu Zhang, and Jian Sun.
\newblock Simple baselines for image restoration.
\newblock In {\em Computer Vision--ECCV 2022: 17th European Conference, Tel
  Aviv, Israel, October 23--27, 2022, Proceedings, Part VII}, pages 17--33.
  Springer, 2022.

\bibitem{hat}
Xiangyu Chen, Xintao Wang, Jiantao Zhou, and Chao Dong.
\newblock Activating more pixels in image super-resolution transformer.
\newblock {\em arXiv preprint arXiv:2205.04437}, 2022.

\bibitem{nafssr}
Xiaojie Chu, Liangyu Chen, and Wenqing Yu.
\newblock Nafssr: stereo image super-resolution using nafnet.
\newblock In {\em Proceedings of the IEEE/CVF Conference on Computer Vision and
  Pattern Recognition}, pages 1239--1248, 2022.

\bibitem{dai2021feedback}
Qinyan Dai, Juncheng Li, Qiaosi Yi, Faming Fang, and Guixu Zhang.
\newblock Feedback network for mutually boosted stereo image super-resolution
  and disparity estimation.
\newblock In {\em Proceedings of the 29th ACM International Conference on
  Multimedia}, pages 1985--1993, 2021.

\bibitem{deng2009imagenet}
Jia Deng, Wei Dong, Richard Socher, Li-Jia Li, Kai Li, and Li Fei-Fei.
\newblock Imagenet: A large-scale hierarchical image database.
\newblock In {\em 2009 IEEE conference on computer vision and pattern
  recognition}, pages 248--255. Ieee, 2009.

\bibitem{dong2015image}
Chao Dong, Chen~Change Loy, Kaiming He, and Xiaoou Tang.
\newblock Image super-resolution using deep convolutional networks.
\newblock {\em IEEE transactions on pattern analysis and machine intelligence},
  38(2):295--307, 2015.

\bibitem{Geiger2012CVPR}
Andreas Geiger, Philip Lenz, and Raquel Urtasun.
\newblock Are we ready for autonomous driving? the kitti vision benchmark
  suite.
\newblock In {\em Conference on Computer Vision and Pattern Recognition
  (CVPR)}, 2012.

\bibitem{gu2019blind}
Jinjin Gu, Hannan Lu, Wangmeng Zuo, and Chao Dong.
\newblock Blind super-resolution with iterative kernel correction.
\newblock In {\em Proceedings of the IEEE conference on computer vision and
  pattern recognition}, pages 1604--1613, 2019.

\bibitem{video2}
Muhammad Haris, Gregory Shakhnarovich, and Norimichi Ukita.
\newblock Recurrent back-projection network for video super-resolution.
\newblock In {\em Proceedings of the IEEE Conference on Computer Vision and
  Pattern Recognition}, pages 3897--3906, 2019.

\bibitem{jeon2018enhancing}
Daniel~S Jeon, Seung-Hwan Baek, Inchang Choi, and Min~H Kim.
\newblock Enhancing the spatial resolution of stereo images using a parallax
  prior.
\newblock In {\em Proceedings of the IEEE conference on computer vision and
  pattern recognition}, pages 1721--1730, 2018.

\bibitem{swinipassr}
Kai Jin, Zeqiang Wei, Angulia Yang, Sha Guo, Mingzhi Gao, Xiuzhuang Zhou, and
  Guodong Guo.
\newblock Swinipassr: Swin transformer based parallax attention network for
  stereo image super-resolution.
\newblock In {\em Proceedings of the IEEE/CVF Conference on Computer Vision and
  Pattern Recognition}, pages 920--929, 2022.

\bibitem{kim2016accurate}
Jiwon Kim, Jung~Kwon Lee, and Kyoung~Mu Lee.
\newblock Accurate image super-resolution using very deep convolutional
  networks.
\newblock In {\em Proceedings of the IEEE conference on computer vision and
  pattern recognition}, pages 1646--1654, 2016.

\bibitem{ledig2017photo}
Christian Ledig, Lucas Theis, Ferenc Huszar, Jose Caballero, Andrew Cunningham,
  Alejandro Acosta, Andrew Aitken, Alykhan Tejani, Johannes Totz, Zehan Wang,
  et~al.
\newblock Photo-realistic single image super-resolution using a generative
  adversarial network.
\newblock In {\em Proceedings of the IEEE conference on computer vision and
  pattern recognition}, pages 4681--4690, 2017.

\bibitem{lei2020deep}
Jianjun Lei, Zhe Zhang, Xiaoting Fan, Bolan Yang, Xinxin Li, Ying Chen, and
  Qingming Huang.
\newblock Deep stereoscopic image super-resolution via interaction module.
\newblock {\em IEEE Transactions on Circuits and Systems for Video Technology},
  31(8):3051--3061, 2020.

\bibitem{mp}
Lei Li, Jingzhu Tang, Ming Chen, Shijie Zhao, Junlin Li, and Li Zhang.
\newblock Multi-patch learning: looking more pixels in the training phase.
\newblock In {\em Computer Vision--ECCV 2022 Workshops: Tel Aviv, Israel,
  October 23--27, 2022, Proceedings, Part II}, pages 549--560. Springer, 2023.

\bibitem{video1}
Sheng Li, Fengxiang He, Bo Du, Lefei Zhang, Yonghao Xu, and Dacheng Tao.
\newblock Fast spatio-temporal residual network for video super-resolution.
\newblock In {\em Proceedings of the IEEE Conference on Computer Vision and
  Pattern Recognition}, pages 10522--10531, 2019.

\bibitem{swinir}
Jingyun Liang, Jiezhang Cao, Guolei Sun, Kai Zhang, Luc Van~Gool, and Radu
  Timofte.
\newblock Swinir: Image restoration using swin transformer.
\newblock In {\em Proceedings of the IEEE/CVF international conference on
  computer vision}, pages 1833--1844, 2021.

\bibitem{lim2017enhanced}
Bee Lim, Sanghyun Son, Heewon Kim, Seungjun Nah, and Kyoung Mu~Lee.
\newblock Enhanced deep residual networks for single image super-resolution.
\newblock In {\em Proceedings of the IEEE conference on computer vision and
  pattern recognition workshops}, pages 136--144, 2017.

\bibitem{10016671}
Jianxin Lin, Lianying Yin, and Yijun Wang.
\newblock Steformer: Efficient stereo image super-resolution with transformer.
\newblock {\em IEEE Transactions on Multimedia}, pages 1--13, 2023.

\bibitem{Menze2015ISA}
Moritz Menze, Christian Heipke, and Andreas Geiger.
\newblock Joint 3d estimation of vehicles and scene flow.
\newblock In {\em ISPRS Workshop on Image Sequence Analysis (ISA)}, 2015.

\bibitem{scharstein2014high}
Daniel Scharstein, Heiko Hirschm{\"u}ller, York Kitajima, Greg Krathwohl, Nera
  Ne{\v{s}}i{\'c}, Xi Wang, and Porter Westling.
\newblock High-resolution stereo datasets with subpixel-accurate ground truth.
\newblock In {\em Pattern Recognition: 36th German Conference, GCPR 2014,
  M{\"u}nster, Germany, September 2-5, 2014, Proceedings 36}, pages 31--42.
  Springer, 2014.

\bibitem{shi2016real}
Wenzhe Shi, Jose Caballero, Ferenc Huszar, Johannes Totz, Andrew~P Aitken, Rob
  Bishop, Daniel Rueckert, and Zehan Wang.
\newblock Real-time single image and video super-resolution using an efficient
  sub-pixel convolutional neural network.
\newblock In {\em Proceedings of the IEEE conference on computer vision and
  pattern recognition}, pages 1874--1883, 2016.

\bibitem{song2020stereoscopic}
Wonil Song, Sungil Choi, Somi Jeong, and Kwanghoon Sohn.
\newblock Stereoscopic image super-resolution with stereo consistent feature.
\newblock In {\em Proceedings of the AAAI Conference on Artificial
  Intelligence}, volume~34, pages 12031--12038, 2020.

\bibitem{timofte2017ntire}
Radu Timofte, Eirikur Agustsson, Luc Van~Gool, Ming-Hsuan Yang, and Lei Zhang.
\newblock Ntire 2017 challenge on single image super-resolution: Methods and
  results.
\newblock In {\em Proceedings of the IEEE conference on computer vision and
  pattern recognition workshops}, pages 114--125, 2017.

\bibitem{timofte2013anchored}
Radu Timofte, Vincent De~Smet, and Luc Van~Gool.
\newblock Anchored neighborhood regression for fast example-based
  super-resolution.
\newblock In {\em Proceedings of the IEEE international conference on computer
  vision}, pages 1920--1927, 2013.

\bibitem{timofte2014a}
Radu Timofte, Vincent De~Smet, and Luc Van~Gool.
\newblock A+: Adjusted anchored neighborhood regression for fast
  super-resolution.
\newblock In {\em Asian conference on computer vision}, pages 111--126.
  Springer, 2014.

\bibitem{vaswani2017attention}
Ashish Vaswani, Noam Shazeer, Niki Parmar, Jakob Uszkoreit, Llion Jones,
  Aidan~N Gomez, Lukasz Kaiser, and Illia Polosukhin.
\newblock Attention is all you need.
\newblock {\em Advances in neural information processing systems}, 30, 2017.

\bibitem{wang2022ntire}
Longguang Wang, Yulan Guo, Yingqian Wang, Juncheng Li, Shuhang Gu, Radu
  Timofte, Liangyu Chen, Xiaojie Chu, Wenqing Yu, Kai Jin, et~al.
\newblock Ntire 2022 challenge on stereo image super-resolution: Methods and
  results.
\newblock In {\em Proceedings of the IEEE/CVF Conference on Computer Vision and
  Pattern Recognition}, pages 906--919, 2022.

\bibitem{Wang2023NTIRE}
Longguang Wang, Yulan Guo, Yingqian Wang, Juncheng Li, Shuhang Gu, Radu
  Timofte, et~al.
\newblock Ntire 2023 challenge on stereo image super-resolution: Methods and
  results.
\newblock In {\em CVPRW}, 2023.

\bibitem{wang2019learning}
Longguang Wang, Yingqian Wang, Zhengfa Liang, Zaiping Lin, Jungang Yang, Wei
  An, and Yulan Guo.
\newblock Learning parallax attention for stereo image super-resolution.
\newblock In {\em Proceedings of the IEEE/CVF Conference on Computer Vision and
  Pattern Recognition}, pages 12250--12259, 2019.

\bibitem{wang2019flickr1024}
Yingqian Wang, Longguang Wang, Jungang Yang, Wei An, and Yulan Guo.
\newblock Flickr1024: A large-scale dataset for stereo image super-resolution.
\newblock In {\em Proceedings of the IEEE/CVF International Conference on
  Computer Vision Workshops}, pages 0--0, 2019.

\bibitem{ipassr}
Yingqian Wang, Xinyi Ying, Longguang Wang, Jungang Yang, Wei An, and Yulan Guo.
\newblock Symmetric parallax attention for stereo image super-resolution.
\newblock In {\em Proceedings of the IEEE/CVF Conference on Computer Vision and
  Pattern Recognition}, pages 766--775, 2021.

\bibitem{yang2008image}
Jianchao Yang, John Wright, Thomas Huang, and Yi Ma.
\newblock Image super-resolution as sparse representation of raw image patches.
\newblock In {\em 2008 IEEE conference on computer vision and pattern
  recognition}, pages 1--8. IEEE, 2008.

\bibitem{remotesensing}
Deniz Yildirim and Ouguz Gungor.
\newblock A novel image fusion method using ikonos satellite images.
\newblock {\em Journal of Geodesy and Geoinformation}, 1(1):75--83, 2012.

\bibitem{ying2020stereo}
Xinyi Ying, Yingqian Wang, Longguang Wang, Weidong Sheng, Wei An, and Yulan
  Guo.
\newblock A stereo attention module for stereo image super-resolution.
\newblock {\em IEEE Signal Processing Letters}, 27:496--500, 2020.

\bibitem{zeyde2010single}
Roman Zeyde, Michael Elad, and Matan Protter.
\newblock On single image scale-up using sparse-representations.
\newblock In {\em International conference on curves and surfaces}, pages
  711--730. Springer, 2010.

\bibitem{zhang2018image}
Yulun Zhang, Kunpeng Li, Kai Li, Lichen Wang, Bineng Zhong, and Yun Fu.
\newblock Image super-resolution using very deep residual channel attention
  networks.
\newblock In {\em Proceedings of the European Conference on Computer Vision
  (ECCV)}, pages 286--301, 2018.

\bibitem{zhang2018residual}
Yulun Zhang, Yapeng Tian, Yu Kong, Bineng Zhong, and Yun Fu.
\newblock Residual dense network for image super-resolution.
\newblock In {\em Proceedings of the IEEE conference on computer vision and
  pattern recognition}, pages 2472--2481, 2018.

\bibitem{fourier}
Man Zhou, Hu Yu, Jie Huang, Feng Zhao, Jinwei Gu, Chen~Change Loy, Deyu Meng,
  and Chongyi Li.
\newblock Deep fourier up-sampling.
\newblock {\em arXiv preprint arXiv:2210.05171}, 2022.

\end{thebibliography}
}

\end{document}